\documentclass[sigconf]{acmart}
\AtBeginDocument{%
  }

\usepackage{amsmath}  
\usepackage{listings}  
\usepackage{xcolor}  
\usepackage{tablefootnote}

\definecolor{lightgray}{gray}{0.9}  
\copyrightyear{2025}
\acmYear{2025}
\setcopyright{acmlicensed}
\acmConference[KDD '25]{Proceedings of the 31st ACM SIGKDD Conference on Knowledge Discovery and Data Mining V.2}{August 3--7, 2025}{Toronto, ON, Canada}
\acmBooktitle{Proceedings of the 31st ACM SIGKDD Conference on Knowledge Discovery and Data Mining V.2 (KDD '25), August 3--7, 2025, Toronto, ON, Canada}
\acmDOI{10.1145/3711896.3737207}
\acmISBN{979-8-4007-1454-2/2025/08}

\begin{document}

\title{Cross-Domain Web Information Extraction at Pinterest}

\author{Michael Farag}
\authornote{These authors contributed equally.}

\email{mfarag@pinterest.com}
\affiliation{%
  \institution{Pinterest}
  \city{Toronto}
  \state{Ontario}
  \country{Canada}
}

\author{Patrick Halina}
\authornotemark[1]
\email{phalina@pinterest.com}
\affiliation{%
  \institution{Pinterest}
  \city{Toronto}
  \state{Ontario}
  \country{Canada}
}

\author{Andrey Zaytsev}
\authornotemark[1]
\email{azaytsev@pinterest.com}
\affiliation{%
  \institution{Pinterest}
  \city{San Francisco}
  \state{California}
  \country{USA}
}

\author{Alekhya Munagala}
\email{amunagala@pinterest.com}
\affiliation{%
  \institution{Pinterest}
  \city{Toronto}
  \state{Ontario}
  \country{Canada}
}

\author{Imtihan Ahmed}
\email{imtihanahmed@pinterest.com}
\affiliation{%
  \institution{Pinterest}
  \city{Toronto}
  \state{Ontario}
  \country{Canada}
}

\author{Junhao Wang}
\email{junhaowang@pinterest.com}
\affiliation{%
  \institution{Pinterest}
  \city{Toronto}
  \state{Ontario}
  \country{Canada}
}

\renewcommand{\shortauthors}{Michael Farag et al.}

\begin{abstract}
The internet offers a massive repository of unstructured information, but it's a significant challenge to convert this into a structured format. At Pinterest, the ability to accurately extract structured product data from e-commerce websites is essential to enhance user experiences and improve content distribution. In this paper, we present Pinterest's system for attribute extraction, which achieves remarkable accuracy and scalability at a manageable cost. Our approach leverages a novel webpage representation that combines structural, visual, and text modalities into a compact form, optimizing it for small model learning. This representation captures each visible HTML node with its text, style and layout information. We show how this allows simple models such as eXtreme Gradient Boosting (XGBoost) to extract attributes more accurately than much more complex Large Language Models (LLMs) such as Generative Pre-trained Transformer (GPT). Our results demonstrate a system that is highly scalable, processing over 1,000 URLs per second, while being 1000 times more cost-effective than the cheapest GPT alternatives.
\end{abstract}



\maketitle

\section{Introduction}
The Internet is a vast source of content and information, but a long-standing challenge is extracting structured data from free form webpages. On Pinterest, users interact with ‘Pins’, which represent content such as products. Understanding structured data attributes from a Pin’s webpage supports many crucial features in the Pinterest application:
\begin{itemize}
\item User Experience: Attributes like the price of a product are displayed as a preview to users within the Pinterest app
\item Distribution: Attributes are used in Pin representations~\cite{baltescu2022itemsage}, which drive recommendation systems~\cite{xia2023transact} and search experiences~\cite{agarwal2024omnisearchsage}
\item Quality: Avoid showing Pins whose website content has substantially changed since Pin creation
\item Website Traffic: The enhanced attributes help drive users to visit the underlying website behind a Pin
\end{itemize}
With more than 500 million monthly active users and 500 billion Pins, Pinterest has developed a system for extracting attributes from webpages that met the challenge of very high accuracy and scalability at a reasonable cost.

\begin{figure}
    \centering
    \fbox{\includegraphics[width=0.5\linewidth]{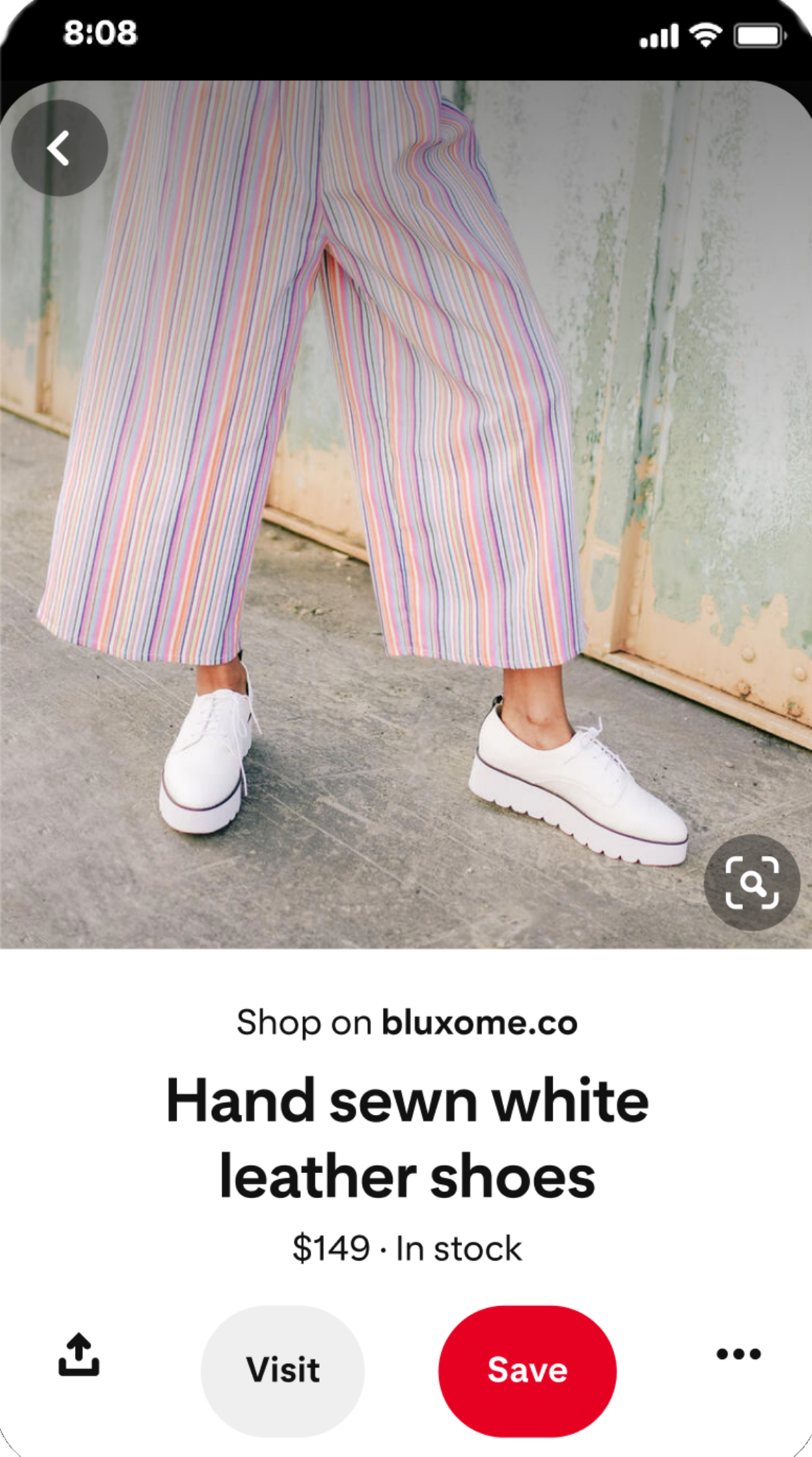}}
    \caption{Pinterest mock product Pin displaying attributes like price and availability}
    \label{fig:enter-label}
\end{figure}

Transforming webpages into structured data is a problem as old as the internet. The ideal approach should achieve high precision and recall for attribute extraction while remaining adaptable to novel website layouts not encountered during training. In addition, Pinterest runs extraction at a large scale, which means the cost per extraction must be as low as possible.  Many ML approaches have been developed, but a universally accurate solution has yet to be found. One reason for the diversity of extraction approaches is the variety of webpage modalities available:
\begin{itemize}
\item Structural: HTML organizes Document Object Model (DOM) nodes in a tree structure. One strategy is to classify the node that contains the desired attribute based on its location in the DOM tree. Classifiers can range from simple regression models to complex Graph Neural Networks (GNNs).
\item Visual: Webpages can be converted into screenshots for analysis by visual models.
\item Text: Text can be extracted from HTML elements or even formatted using open source libaries like Inscriptis~\cite{Weichselbraun2021}. The raw text of a page can be processed by Large Language Models (LLMs) or Natural Language Processing (NLP) solutions.
\end{itemize}

\begin{figure}
    \centering
    \includegraphics[width=0.5\linewidth]{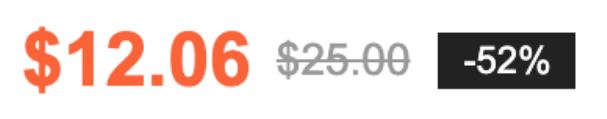}
    \caption{An example of how a sale price is differentiate from the original list price using a visual cue: a strikethrough style.}
    \label{fig:price}
\end{figure}

These modalities each have challenges. Webpages share a common visual design language that humans understand. A simple example is sale pricing: to indicate a product is on sale, it's common to display the original (higher) list price beside the final sale price, with a strike through the original list price, as seen in figure~\ref{fig:price}. When representing the webpage as pure text, this strike style gets lost. HTML structure has many ways to represent a strike through, which can change between domains, so it requires the extractor to read javascript and CSS, then correctly predict rendering engine rules to know whether text is rendered with a strikethrough. Screenshots contain visual patterns, but their larger storage also requires larger models to understand. They also lack key non-visual information like the URLs of links or images.

To deal with these problems, we introduce Visual Page Representation (VPR), a novel representation of a webpage that’s more compact than HTML yet expressive enough for small models to learn cross website extraction patterns. 

This representation works by saving every visible HTML node containing text or images, along with pertinent attributes like URL links and styles (colors, text size, etc). Our system can use a simple eXtreme Gradient Boosting (XGBoost)~\footnote{https://xgboost.readthedocs.io/} model to classify whether a webpage represents a product, error page or some other content type. For content types like products, we then feed the webpage representation into another XGBoost model to classify which elements represent each attribute of interest (eg. price, title, main image.) The richness of our novel representation allows simple models to have extremely high accuracy while running at low cost on CPU hardware.

With our novel representation and accurate low cost models, this left rendering as the most expensive part of the pipeline. Compared to simply requesting the static HTML from a URL, visual rendering incurs higher computational costs in our system as well as the website server (since it must provide images, CSS and execute dynamic javascript requests.) After deployment we minimized this costly step by distilling our accurate cross-website models into static HTML only models for specific domains.

To summarize, we discuss Pinterest's web information extraction system which has proven:
\begin{enumerate}
    \item Scalable: At the time of writing we process over 1,000 URLs per second
    \item Accurate: Our offline tests show extraction surpasses LLM alternatives
    \item Affordable: Our system is orders of magnitude cheaper than more complex LLM solutions.
\end{enumerate}

\section{Related Work}
A variety of methods exist to extract structured data from webpages.

For common use cases like titles, main images and prices, open web standards~\footnote{https://ogp.me/}\footnote{https://schema.org/} exist to embed metadata within the HTML. Our experience shows that these data sources are often unreliable and inaccurate (see schema.org and Open Graph results in Table \ref{tab:results}). Many pages are missing this data or the embedded metadata contradicts what users see on the webpage. Simple approaches like regex based text matching are also unreliable and require fragile logic specific to each website. To deal with these issues, various ML approaches have been developed.

An early and popular ML method for structured attribute extraction is Wrapper Induction (WI)~\cite{Kushmerick00,LiuML15,FerraraMFB14,SchulzLG16,SleimanC13}, developed in 1999. This method uses a linear model to identify HTML nodes associated with an attribute. It is fast to train, only requires static HTML for input features (reducing the rendering times compared to visual approaches), and inference requires little computation. However, it requires training a model for each specific website layout. This limits scalability to new websites and makes it susceptible to sudden breakages when a website updates its layout. It makes sense that each website requires its own model, as Wrapper Induction is meant to learn templates, which aren’t shared across domains. Pinterest has successfully used this method for extracting product information from merchant websites. However, scaling beyond 10,000 websites was challenging as the effort to maintain existing models required more and more human labelling capacity. This prompted us to find an extraction approach that can generalize across unseen websites.

There have been a slew of modern approaches to structured web extraction which use Deep Neural Networks (DNNs). MarkupLM~\cite{li2021markuplm} is a BERT like transformer architecture which is specialized for HTML. It has been trained to understand the semantics of an HTML node’s text content plus its structural relationship within the DOM tree. However, MarkupLM's precision and recall were less than our proposed solution with a 10x higher cost. 
SimpDOM~\cite{zhou2021simplified} is another model that combines structure with text understanding. Approaches like 
DocFormerV2~\cite{appalaraju2024docformerv2} and LayoutLMV3~\cite{huang2022layoutlmv3} work on visual representations of documents. While these are generalized to any visual document, they can be applied to structured webpage extraction too. For the scale of our system, we not only require very high extraction precision, but low cost in computation. These DNN models are too expensive to run on each webpage. The cost comes from the number of parameters in a model as well as the input size required. We’ve observed that the median number of tokens in HTML is about 20k tokens. Smaller architectures like BERT have context windows of only 512 tokens. This requires strategies to partition a webpage and feed it into the model piece by piece as well as reconciling extracted values from all the partitions.

Finally, the emergence of LLMs like GPT provide zero-shot methods for extraction. By simply providing prompts along with HTML or text, GPT can extract structured data in JSON format. New variants combine LLMs with visual models, allowing rendered screenshot images to be provided as well. The downside of these approaches is the cost. For research purposes, we benchmarked various GPT models. Cheaper priced models (eg. GPT 4o-mini) didn't have high enough precision for our needs. More expensive models like GPT-o1 increase accuracy, but come with an order of magnitude increase in price. Unfortunately, even the cheapest GPT models are several orders of magnitude more expensive than our proposed solution.

\section{System Overview}
\begin{figure*}
    \centering
    \includegraphics[scale=0.5]{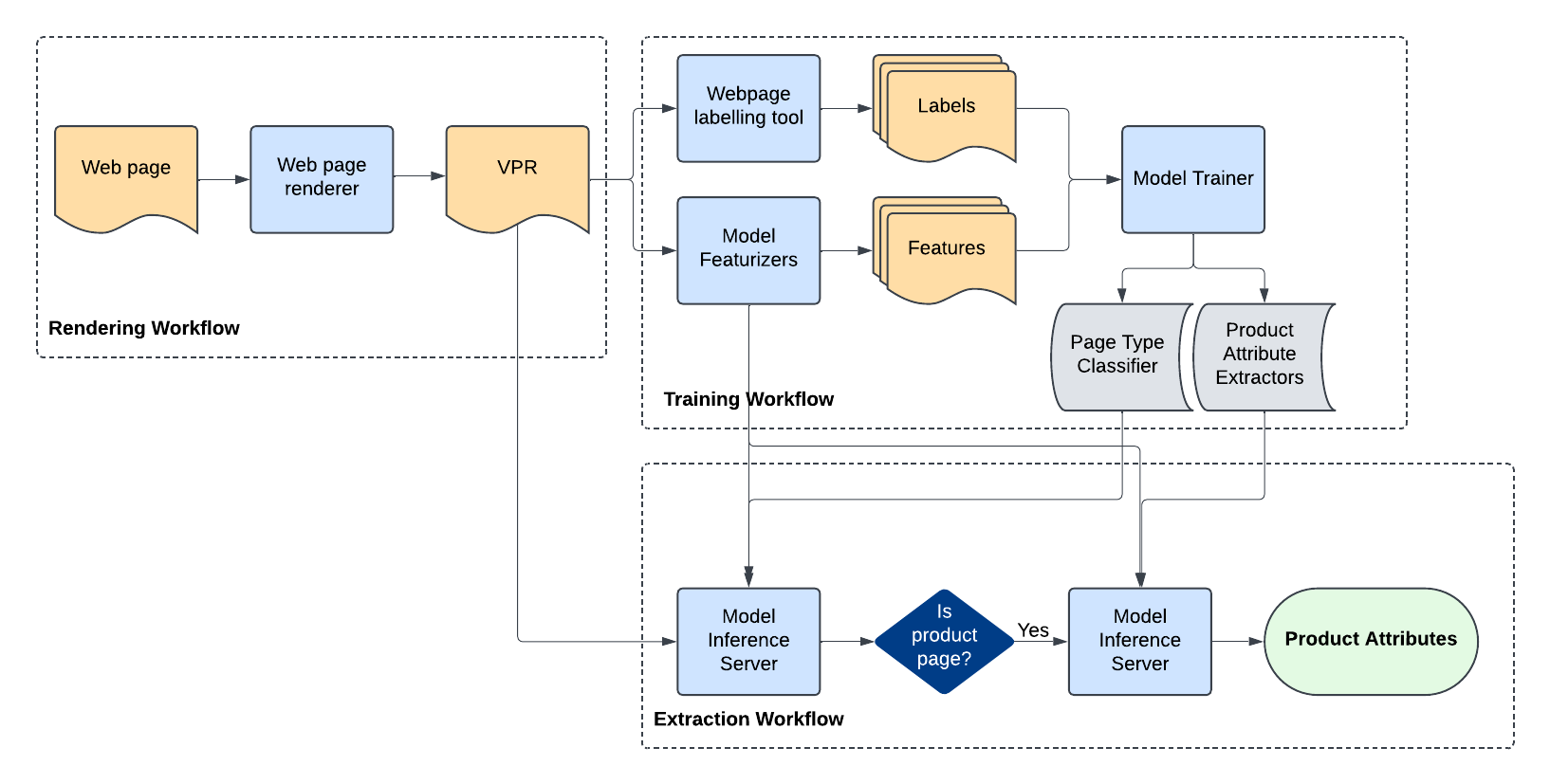}
    \caption{System Architecture}
    \Description{System Architecture}
    \label{fig:system_architecture}
\end{figure*}

The system architecture comprises three primary workflows as shown in figure~\ref{fig:system_architecture}: Rendering, Training, and Extraction.

In the \textit{Rendering Workflow}, a webpage URL is fed into a \textit{Web Page Renderer}, which captures the HTML and visual elements, creating a \textit{Visual Page Representation (VPR)}.

For the \textit{Training Workflow}, the VPR is used by a \textit{Webpage Labelling Tool} where human annotators generate structured labels. 
\textit{Model Featurizers} extract relevant features from the VPR, which are then used to train models in the \textit{Model Trainer}, including a \textit{Page Type Classifier} for identifying product pages and a \textit{Product Attributes Extractor} for detailed attribute extraction. The feauturization logic and trained models are deployed to the \textit{Model Inference Library}

The \textit{Extraction Workflow} uses the \textit{Page Type Classifier} to determine a VPR's page type. For product pages, the system uses the \textit{Product Attributes Extractor} to identify and extract attributes, outputting structured product information.

This architecture efficiently aligns rendering, training, and extraction processes to deliver accurate product data.

\section{Labelling Tool}
\begin{figure*}
    \centering
    \includegraphics[width=0.9\linewidth]{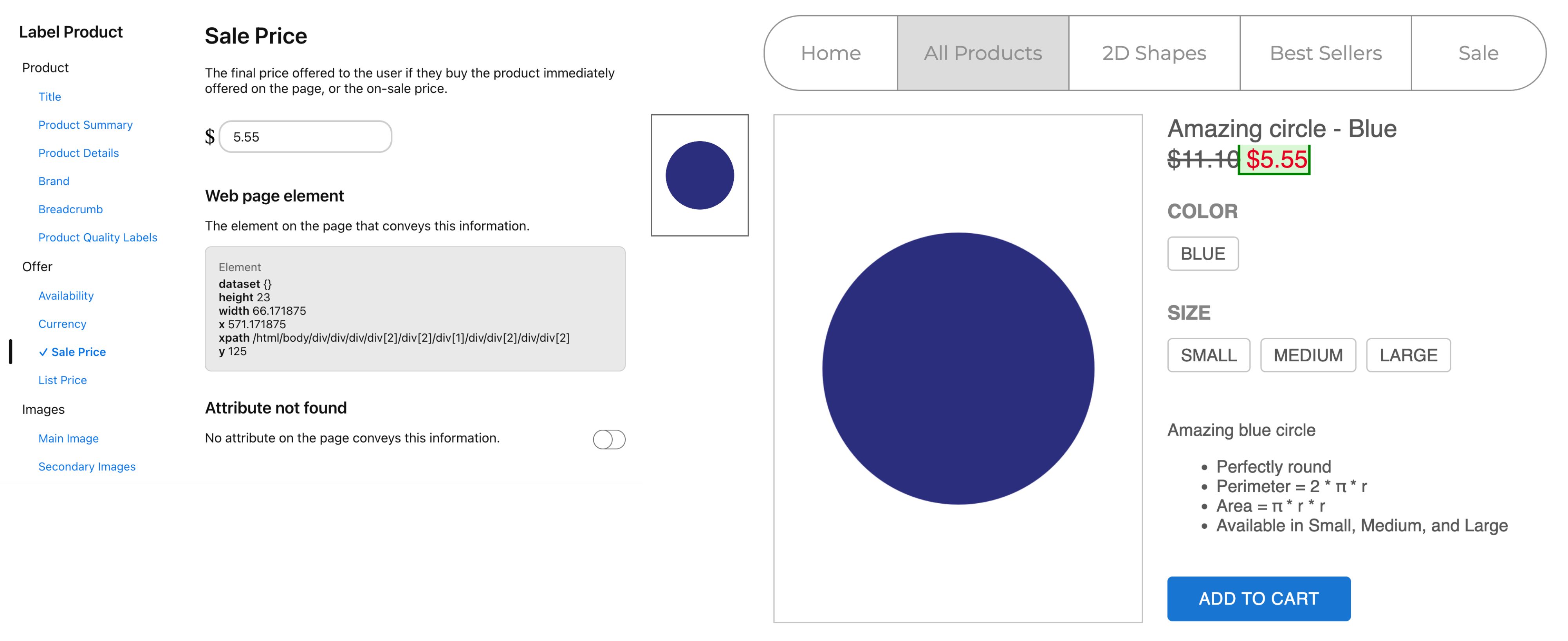}
    \caption{Our webpage labelling tool, with VPR rendered page on the right. Visual cues like strike through on text and text size are maintained in VPR.}
    \label{fig:web_labelling_tool}
\end{figure*}

In order to train models for our new VPR page modality, we need labels of VPR elements that represent attributes. We built our own webpage labelling tool since existing tools won't work with our proprietary representation. The labelling tool has a task management system where webpages are preloaded for labelling. The tool then serves tasks to our labelling team. A representation of the page is shown in one window, while questions about attributes to label are shown in another window. Figure \ref{fig:web_labelling_tool} shows a screenshot of the UI. For each attribute, labellers select the element from the webpage containing the attribute's data, then enter the final value. Some attributes have preset values to select from (eg. in stock, out of stock, pre order) while others are free form text. If no element on the page directly contains the value of the attribute, labellers can simply enter the value directly. This has allowed us to collect tens of thousands of labels. We use it to create datasets as well as audit the metadata quality of our live system.

\section{Visual Page Representation}
The Visual Page Representation (VPR) is a novel approach that bridges the gap between the underlying HTML structure of a webpage and its rendered visual layout. This representation provides a comprehensive view of web content by aligning and unifying the visual attributes and spatial geometry with the HTML elements, thereby facilitating advanced applications such as product attribute extraction.

VPR offers several advantages over traditional methods that rely solely on HTML or page screenshots for webpage analysis:

\textbf{Comprehensive Contextual Understanding:} While HTML provides the structural backbone of a webpage, it lacks direct information about how elements are visually rendered. Purely HTML-based analyses may overlook or misinterpret the spatial and visual relationships between elements, such as overlapping content or style-induced changes. By incorporating spatial coordinates (e.g. relative position of title and main image) and visual properties (e.g. text size and strikethrough), VPR captures the exact rendering of a webpage at a specific moment, enabling precise understanding of content layout and appearance. This is particularly beneficial for tasks that require nuanced layout understanding, such as detecting prominent features like banners or highlights.

\textbf{Enhanced Spatial Information:} Screenshots provide a visual snapshot but lack the underlying semantic relationships and hierarchies present in the DOM. This often makes it difficult to link visual elements back to their functional roles or extract interaction points. Eg. screenshots don't contain the URL sources of images. VPR combines the strengths of screenshots and HTML by embedding visual data within the DOM context. This dual-layered information allows for more accurate mapping of visual elements to their semantic roles, essential for applications like attribute extraction where understanding both element placement and function is crucial.

\subsection{VPR Generation}
VPR is generated by a Pinterest developed rendering service based on Chromium~\footnote{https://www.chromium.org/} browser, which uses the Blink~\footnote{https://www.chromium.org/blink/} rendering engine. For a given URL, the service renders the page and generates VPR in a headless browser. In general, rendering engines download assets like HTML, CSS, JS, and images from a web server to determine the bitmap to display on a screen. This is a multistage process that involves HTML parsing, building DOM tree, building layout tree, executing Javascript and rendering. During this process Blink saves intermediate structures in memory. These intermediate structures are captured and used for generating VPR according to the schema mentioned in section~\ref{sec:vpr_schema}.

\subsection{VPR Schema Structure}
\label{sec:vpr_schema}
The VPR schema is structured as a JSON object that encapsulates several core components of a webpage: 

\begin{lstlisting}  
{  
  "url": <string>, 
  "title": <string>,
  "width": <integer>,
  "height": <integer>,
  "imageElements": [<ImageElement>, ...],  
  "textElements": [<TextElement>, ...],  
  "actionElements": [<ActionElement>, ...],  
  "xpathTree": [<XPathNode>, ...],  
  "version": <string>
}  
\end{lstlisting} 

\textbf{Image Elements:}    
Each image element on the webpage is characterized by:    
\begin{itemize}    
    \item \texttt{x, y}: Coordinates of the top-left corner of the image.  
    \item \texttt{width, height}: Dimensions in pixels.  
    \item \texttt{xpathId}: Identifier linking the image to its corresponding DOM node.  
    \item \texttt{src}: URL of the image source.  
\end{itemize}    
  
\textbf{Text Elements:}    
Each text element on the webpage is characterized by:    
\begin{itemize}    
    \item \texttt{x, y}: Coordinates of the top-left corner of the text.  
    \item \texttt{width, height}: Dimensions in pixels.  
    \item \texttt{xpathId}: Identifier linking the text to its corresponding DOM node.  
    \item \texttt{fontSize}: Size of the text font.  
    \item \texttt{lineThrough}: Boolean indicating if the text has a strike through style.  
    \item \texttt{text}: Actual text content of the node.  
\end{itemize}    
  
\textbf{Action Elements:}  
Each action element on the webpage represents an interactive part of the webpage, such as a link or button:  
\begin{itemize}  
    \item \texttt{x, y}: Coordinates of the top-left corner of the action point. 
    \item \texttt{width, height}: Dimensions in pixels.  
    \item \texttt{xpathId}: Identifier linking the action to the corresponding DOM node.  
    \item \texttt{href} (optional): URL if the element is a hyperlink.  
\end{itemize}  
  
\textbf{XPath Tree:}    
This component models the hierarchical structure of the DOM:    
\begin{itemize}    
    \item \texttt{tagName}: HTML tag associated with the node.  
    \item \texttt{parentId}: Identifier of the node's parent in the DOM tree.  
    \item \texttt{xpathId}: Unique identifier for nodes linked to visible elements.  
\end{itemize}

\section{Page Type Classifier}
For the production Page Type classifier, we focused on accurately categorizing various webpage types such as PRODUCT, SOFT404 and JUNK. Our approach leveraged a set of refined features, including web-structural features based on VPRs such as image text ratio, as well as content-level features such as title and textual embeddings. We train a robust XGBoost model designed to maintain performance across different classes while improving precision where most needed. A key lesson learned during the development process was that reducing the PRODUCT recall slightly below 99.9\% led to significant gains in precision without causing substantial regressions in other metrics. This strategy reduced outdated or irrelevant PRODUCT false positives and halved the false positives for PRODUCT. By enhancing the clarity between PRODUCT and other classes, we saw a broader 10\% PRODUCT precision improvement and over 80\% reduction in misclassifications of SOFT404 as PRODUCT.

\section{Product Metadata Extractors}
In this paper, we focus on five primary attributes found in product pages: title, currency, sale price, list price, and main image. We have developed three independent extractors for these attributes: a title extractor, a main image extractor, and a price extractor, which handles the extraction of currency, sale price, and list price. Using VPR, we engineered numerous features for each attribute and trained XGBoost models. These models classify the elements within the VPR to determine whether they represent a specific attribute. The modeling process involves a series of steps: candidate selection, feature engineering, and classification. These steps are detailed in the next section.
\subsection{Modeling}
\subsubsection{Problem Formulation}  
Given a set of webpages spanning a diverse range of domains $\mathcal{D} = \{W_1, W_2, \ldots, W_n\}$, each webpage $W_i$ is represented by a VPR $V_i$.   
  
The VPR encapsulates visual and structural features, consisting of elements $E_i = \{E_{i1}, E_{i2}, \ldots, E_{im}\}$. Each element corresponds to a segment of the rendered webpage, incorporating both visual layout and semantic data.  
  
For each webpage $W_i$, there are multiple attributes of interest $\mathcal{L} = \{A_1, A_2, \ldots, A_k\}$.   
  
Each attribute $A_j$ has a corresponding element $E_{ij} \in E_i$ and an associated value $V_{ij}$ that have been manually labeled by humans:      
$L(W_i, A_j) = (E_{ij}, V_{ij})$

The task is to learn a function $f: \mathcal{D} \times \mathcal{L} \to \{(E, V)\}$, such that for a given webpage $W_i$ and attribute $A_j$, the function predicts the correct element and value.

\subsubsection{Candidates Selection}
The initial step in the modeling process is candidate selection. As presented in the VPR schema section, each page is represented by a list of text, action, and image elements. In this step, we apply heuristics to identify which VPR elements might contain a specific attribute. For the price models, we utilize pattern matching to identify text elements that contain currency symbols and strings resembling prices. For the title model, all text elements from the VPR are selected as potential candidates. For the main image model, all image elements are selected as candidates.

\subsubsection{Feature Engineering}
We engineered numerous features to feed as input to the XGBoost models. These features encapsulate different aspects of an element. While some features are shared between the extractors, others are specific to a particular extractor and do not apply to other extractor types. The features can be broadly arranged into the following categories:

\textbf{Layout-based features:}
capture the spatial arrangement and positioning of elements on the product page. They are crucial for understanding how elements relate to each other visually. Examples of layout based features include x-coordinate, y-coordinate, width, height, area, distance to largest image, distance to largest text, number of candidates in the same row/column, etc.

\textbf{Style-based features:}
capture the visual characteristics of text elements. These include features such as font size and whether the text has a strike through it.

\textbf{Rank-based features:}
prioritize and rank elements within their context, indicating the relative position of the element’s feature within the candidate list. Examples of these features include width, height, area ranks, as well as distance from largest image, and largest text ranks, etc.

\textbf{Attribute-specific features:}
These features are unique to the attribute being extracted and provide little value for other attributes. Examples include whether an element is lazy-loaded or clickable for the main image model; the number of elements with the same text value for the title model; and price rank or the number of elements with the same price for price models.

\subsubsection{Classification}
In the classification step, we extract features for each candidate and utilize trained XGBoost models to classify the candidate. For the title and main image models, we train a binary classifier to determine whether an element contains the respective attribute. For the price model, we train a multi-class classifier to detect list price and sale price. Currency is determined from the extracted list and sale price values using string matching on different currency symbols.

Note that the models output a score for each candidate. We also perform threshold tuning to identify the cut-off scores for each attribute. If the highest scoring candidate's score is below the set threshold, the classification result is returned as empty.

\subsection{Distilling to Wrapper Induction}
To reduce rendering costs, we distill XGBoost models to Wrapper Induction (WI) models on some domains. WI models operate solely on HTML, which is cheaper to generate than VPR since it doesn't require browser rendering. However, WI models traditionally require labeled data for each domain, which typically involves labor-intensive manual labeling.

To address the challenge of labeling data for WI models, we used XGBoost to generate accurate cross-domain predictions from VPR data. By mapping these predictions to static HTML DOM nodes where possible, they served as automated labels, replacing the need for manual involvement. This approach enabled us to train WI models with accuracy comparable to VPR-based extractors but at significantly lower computational costs.

In domains where WI models matched the accuracy of XGBoost, we switched entirely to WI models, reducing the need for costly visual rendering. This shift to static HTML processing significantly reduced operational costs and enhanced the scalability of our data extraction efforts, ultimately enabling us to transition approximately 60\% of domains to a more cost-effective approach.

\section{Dataset}
We created a dataset designed to benchmark the extraction of product attributes, featuring a diverse collection of webpage representations. The dataset includes labels for all core website modalities: HTML, text, screenshots, as well as our novel VPR.
Our dataset is structured with distinct domains for training and testing to ensure generalizability. This setup guarantees that no test domain is seen during training, allowing us to rigorously evaluate the system's ability to generalize to new, unseen domains. 
The dataset splits stats are shown in Table~\ref{tab:dataset_stats} and the coverage of labeled attributes are shown in Table~\ref{tab:attributes_coverage}.

The dataset was constructed by sampling URLs from 2,340 distinct domains, resulting in a total of 3,322 webpages. Within each domain, we aimed to stratify our samples by including one URL featuring an in-stock product and another featuring an out-of-stock product, wherever feasible. The availability of individual URLs was initially inferred from labels generated by our existing production pipeline. 

After collecting the URLs, we simultaneously gathered the HTML, VPR, and screenshots from each webpage. This synchronous collection was accomplished using our internal web rendering system, allowing for consistent snapshots of each webpage. 
A trained team of internal labellers then labeled each attribute using our webpage labelling tool.

For each product webpage, labellers were tasked with identifying and recording various relevant product attributes. Typically, the task involves finding textual elements corresponding to specific attributes, such as sale price and product title. By clicking on the VPR element that corresponds to these attributes, the text value, the XPath, and bounding boxes of the relevant node are captured and recorded.

Our labellers were trained using a comprehensive instructional guide developed by our engineers and product managers, which includes detailed examples and a focus on handling tricky edge cases in labelling each attribute.

\begin{table}
    \centering
    \caption{Dataset Stats}
    \label{tab:dataset_stats}
    \begin{tabular}{lccc}
        \toprule
         & Training & Test & Total\\
        \midrule
        Number of Domains & 1325 & 1015 & 2340\\
        Number of Pages   & 2105 & 1217 & 3322\\
        \bottomrule
    \end{tabular}
\end{table}

\begin{table}
    \centering
    \caption{Coverage of Labeled Attributes}
    \label{tab:attributes_coverage}
    \begin{tabular}{cc}
        \toprule
        Attribute & Coverage\\
        \midrule
        Title & 0.9954\\
        Description & 0.7157\\
        Main Image & 0.9717\\
        Availability & 0.9550\\
        Currency & 0.9691\\
        Sale Price & 0.9320\\
        List Price&0.2130\\
        \bottomrule
    \end{tabular}
\end{table}

\section{Experiments and Results}
Our experiments evaluated various extraction methods using HTML, text, screenshots, and our proposed VPR. 
The evaluation of different extractors across webpage modalities reveals significant insights into their performance, trade-offs, and operational costs. 
Table~\ref{tab:results} highlights these findings.

\begin{table*}[tb]  
\centering      
\caption{Extractors Results, P and R denotes Precision and Recall, respectively.}
\label{tab:results}      
\begin{tabular}{@{}llccccccccccc@{}}      
\hline      
\textbf{Webpage Modality} & \textbf{Extractor} & \multicolumn{2}{c}{\textbf{Title}} & \multicolumn{2}{c}{\textbf{Main Image}} & \multicolumn{2}{c}{\textbf{Currency}} & \multicolumn{2}{c}{\textbf{Sale Price}} & \multicolumn{2}{c}{\textbf{List Price}} \\ \hline      
 & & \textbf{P} & \textbf{R} & \textbf{P} & \textbf{R} & \textbf{P} & \textbf{R} & \textbf{P} & \textbf{R} & \textbf{P} & \textbf{R} & \textbf{} \\ \hline      

HTML & schema.org & 0.9485 & 0.8732 & 0.6636 & 0.7663 & 0.9237 & 0.8318 & 0.8995 & 0.8101 & - & - \\ 
HTML & Open Graph & 0.8892 & 0.7911 & 0.6594 & 0.7306 & 0.9369 & 0.7253 & 0.8930 & 0.7163 & 0.8125 & 0.0522 \\ 

HTML & MarkupLM & 0.9055 & 0.9945 & - & - & 0.9500 & 0.8720 & 0.9290 & 0.9566 & 0.7829 & 0.8211 \\

Text & GPT 4o-mini & 0.9792 & 0.9899 & - & - & 0.9416 & 0.9886 & 0.9381 & 0.9875 & 0.7566 & 0.9497 \\ 
Text + Screenshot & GPT 4o-mini & \textbf{0.9942} & 0.9983 & - & - & 0.9531 & 0.9904 & 0.9546 & 0.9973 & 0.8581 & 0.8581 \\ 
VPR & GPT 4o-mini & 0.8544 & 0.9923 & - & - & 0.9534 & \textbf{0.9965} & 0.9550 & 0.9903 & 0.8095 & 0.9714 \\ 

Text & GPT 4o & 0.9847 & 0.9716 & - & - & 0.9479 & 0.9694 & 0.9506 & 0.9699 & 0.8566 & 0.9549 \\ 
Text + Screenshot & GPT 4o & \textbf{0.9942} & 0.9991 & - & - & \textbf{0.9540} & 0.9904 & 0.9546 & 0.9982 & 0.8754 & 0.9835 \\ 
VPR & GPT 4o & 0.9240 & 0.9955 & - & - & 0.9533 & 0.9939 & 0.9626 & 0.9921 & 0.8847 & 0.9754 \\     

Text & GPT o1 & 0.9917 & 0.9975 & - & - & 0.9476 & 0.9930 & 0.9487 & 0.9964 & 0.8608 & 0.9791 \\ 
Text + Screenshot & GPT o1 & \textbf{0.9942} & 0.9983 & - & - & 0.9531 & 0.9904 & 0.9546 & 0.9973 & 0.8581 & 0.9711 \\ 
VPR & GPT o1  & 0.9679 & \textbf{1.0} & - & - & 0.9477 & 0.9956 & 0.9564 & \textbf{1.0} & 0.9053 & 0.9795 \\

VPR & XGBoost & 0.9603 & 0.9991 & \textbf{0.9804} & \textbf{0.9469} & 0.9446 & 0.9629 & \textbf{0.9726} & \textbf{1.0} & \textbf{0.9652} & \textbf{1.0} \\ \hline

\end{tabular}      
\end{table*}  

\subsection{Experimental Setup}
In our approach to extracting product metadata from webpage content using GPT models, we deploy system prompts that are specifically tailored to the demands of different models and input modalities. We utilize multimodal prompts that combine both textual and visual information, allowing the model to harness both sources in its analysis. Additionally, we use text-only prompts to focus exclusively on the textual content and VPR-only prompts to understand and interpret structured text elements based on their spatial layout on a webpage. 
The Structured Outputs API~\footnote{https://platform.openai.com/docs/guides/structured-outputs} with Pydantic schemas ensures data adheres to predefined formats. For the o1 model, we rely on its capabilities without step-by-step instructions, while for 4o and 4o mini models, we enhance reasoning by prompting them to work methodically through tasks. This approach maximizes each model's potential across various input scenarios.

For MarkupLM, we fine-tuned it to perform multi-class classification for extracting text-based attributes like title, sale price, and list price from webpages. MarkupLM is trained on both text and the XPath of HTML nodes. Because its context length is limited to 512, we divide each webpage into multiple 512-token chunks for both training and inference. During inference, we then reconcile the final values by selecting the node with the highest score for each attribute across all chunks and using that as the final output for the page. While the model itself extracts the title, sale price, and list price, we further apply regex-based rules on the extracted sale and list prices in order to determine currency.

\subsection{HTML-Based Extraction}

The results for HTML extractors using schema.org and Open Graph have the poorest performance. This highlights the need for advanced extraction approaches from web content to achieve better accuracy and reliability. On the other hand, while MarkupLM has good precision in identifying relevant elements, it usually have lower recall.

\subsection{Text-Based Extraction with GPT}

Text-based models such as GPT variants exhibit strong performance in textual attribute extraction, particularly for titles and currency. Precision and recall scores are notably high, with GPT o1 achieving close to perfect recall.

\paragraph{Enhancement with Screenshots:} Incorporating screenshots alongside text further boosts performance. The \texttt{Text + Screenshot} approach achieves higher precision and recall than the text-only case, emphasizing how visual context complements and enriches textual analysis. 

\paragraph{Main Image Limitations:}
Text and screenshots lack direct access to image URLs, essential for accurate main image extraction. Thus, structured data sources like HTML or VPR are necessary as they include these URLs.

\subsection{Impact of Visual Page Representation}
VPR emerges as a cost-effective and efficient alternative, retaining essential visual cues without the overhead of full screenshots.

\paragraph{GPT with VPR:} GPT o1 worked surprisingly well using VPR as an input, which GPT was not previously exposed to. Its extraction performance closely matched and sometimes exceeded GPT models that used text and screenshots as inputs. This is demonstrated by list price, where precision was 0.905 compared to 0.858. VPR is also much more compact than screenshots and text, leading to a significant reduction in storage costs as clear from Table~\ref{tab:webpage_modalities}.

\begin{table}[H]
    \centering
    \caption{Webpage Modalities Data Size}
    \label{tab:webpage_modalities}
    \begin{tabular}{lc}
        \toprule
        Webpage Modality & Mean Data Size (KB)\\
        \midrule
        HTML & 477.393\\
        Text & 18.6458\\
        Screenshot & 566.537\\
        VPR & 32.26\\
          \bottomrule
    \end{tabular}
\end{table}

\paragraph{VPR with XGBoost:} This combination demonstrates exceptional performance, especially in extracting sale and list prices as well as main images, achieving perfect recall for these price attributes and higher precision compared to GPT. While XGBoost is slightly less precise in extracting titles, the alternative titles it generates are usually representative of the content. Post deployment experience has shown that title discrepancies are rarely significant enough to impact the user experience.
Currency extraction highlights a key difference between XGBoost and GPT. XGBoost struggles with ambiguous symbols like the dollar sign, which might denote USD, CAD, or AUD. In contrast, GPT effectively resolves this by using contextual clues, such as recognizing a Canadian address to infer CAD. This contextual inference gives GPT an advantage in managing currency ambiguities.
Overall, VPR with XGBoost proves to be more cost-effective than GPT alternatives, offering an efficient and scalable solution for web data extraction.

\subsection{Key Trade-offs and Insights}  

\paragraph{Cost and Performance Balance:} While GPT-based models offer high accuracy, their computational and resource costs are significantly higher compared to using XGBoost with VPR. This illustrates a clear trade-off between accuracy and cost, as shown in Tables~\ref{tab:webpage_modalities} and~\ref{tab:cost_analysis} where XGBoost extractors are 1000 times less costly compared to the cheapest GPT alternative (GPT 4o-mini).
To estimate GPT costs, we calculate the total number of tokens processed during inference on the test dataset and utilize the available public pricing rates.\footnote{The prices reported are as of February 2025.}

\begin{table}[H]
    \centering
    \caption{Average Extraction Cost per Webpage}
    \label{tab:cost_analysis}
    \begin{tabular}{llc}
        \toprule
        Webpage Modality & Extractor & Cost (USD)\\
        \midrule
        HTML & MarkupLM & \$0.000069\\
        Text & GPT 4o-mini & \$0.002027\\
        Text + Screenshot & GPT 4o-mini & \$0.002524\\
        VPR & GPT 4o-mini & \$0.003676\\
        Text & GPT 4o & \$0.016329\\
        Text + Screenshot & GPT 4o & \$0.015523\\
        VPR & GPT 4o & \$0.043682\\
        Text & GPT o1 & \$0.225634\\
        Text + Screenshot & GPT o1 & \$0.203921\\
        VPR & GPT o1 & \$0.386073\\
        VPR & XGBoost & \$0.0000079\\
        \bottomrule
    \end{tabular}
\end{table}

\paragraph{Effectiveness of VPR:} VPR not only closely matches but also surpasses the accuracy achieved by GPT extractors when processing certain attributes, offering an efficient and viable alternative. By successfully integrating the visual layout with HTML structure, VPR represents a novel approach that maintains high performance while efficiently handling varied web content.

\subsection{Post Deployment Results}  
Initially deployed in production across more than 8,000 websites, the system achieved a post-deployment engagement weighted average precision of 98\% across main image, title, availability, sale price, list price and description. The system successfully scaled beyond 1,000 URLs per second, at an average cost of \$0.0079 to process 1,000 URLs, including rendering, featurization and post processing.

\section{Conclusion}

In summary, we have implemented a scalable and efficient system for extracting structured product data from webpages at Pinterest, leveraging a novel webpage representation: VPR. This approach seamlessly integrates the visual layout with HTML structure, enabling precise attribute extraction using cost-effective models like XGBoost.

Our system successfully processes thousands of URLs per second, significantly reducing operational costs compared to complex models such as large language models. By automating the creation of Wrapper Induction models, we further decrease visual rendering expenses, enhancing system efficiency for both dynamic and static HTML pages.

This system's deployment at Pinterest highlights the transformative potential of VPR, delivering scalable, accurate, and cost-effective solutions that align with the demands of high-volume web extractions.


\newpage
\bibliographystyle{ACM-Reference-Format}
\bibliography{references}

\end{document}